# Detection and classification of masses in mammographic images in a multi-kernel approach

Sidney M. L de Lima[a], Abel G. da Silva-Filho[a], and Wellington Pinheiro dos Santos[b]

[a]*Center of Informatics - CIn, Federal University of Pernambuco, UFPE - Recife, Brazil {smll,agsf}@cin.ufpe.br*
[b]*Department of Biomedical Engineering, Federal University of Pernambuco, UFPE - Recife, Brazil wellington.santos@ufpe.br*

ARTICLE INFO

*Article history:*
Received 00 December 00
Received in revised form 00 January 00
Accepted 00 February 00

Keywords:
Breast cancer
Mammography
Multi-resolution wavelets
Extreme learning machines
Support vector machines

ABSTRACT

According to the World Health Organization, breast cancer is the main cause of cancer death among adult women in the world. Although breast cancer occurs indiscriminately in countries with several degrees of social and economic development, among developing and underdevelopment countries mortality rates are still high, due to low availability of early detection technologies. From the clinical point of view, mammography is still the most effective diagnostic technology, given the wide diffusion of the use and interpretation of these images. Herein this work we propose a method to detect and classify mammographic lesions using the regions of interest of images. Our proposal consists in decomposing each image using multi-resolution wavelets. Zernike moments are extracted from each wavelet component. Using this approach we can combine both texture and shape features, which can be applied both to the detection and classification of mammary lesions. We used 355 images of fatty breast tissue of IRMA database, with 233 normal instances (no lesion), 72 benign, and 83 malignant cases. Classification was performed by using SVM and ELM networks with modified kernels, in order to optimize accuracy rates, reaching 94.11%. Considering both accuracy rates and training times, we defined the ration between average percentage accuracy and average training time in a reverse order. Our proposal was 50 times higher than the ratio obtained using the best method of the state-of-the-art. As our proposed model can combine high accuracy rate with low learning time, whenever a new data is received, our work will be able to save a lot of time, hours, in learning process in relation to the best method of the state-of-the-art.

## 1. Introduction

Breast cancer is the leading cause of death of women around the world, both in developed and underdevelopment countries [1]. According to the World Health organization (WHO), in 2012, about 1.7 million new cases of breast cancer emerged in the world [1]. Additionally, breast cancer is the most common type of cancer in 140 countries of a total of 182 evaluated nations [1]. The incidence of breast cancer increased 20% between the years 2008 and 2012, as well as mortality rates augmented by around 14% [1].

According to Brazil's National Institute of Cancer, approximately 30% of cases of breast cancer could be prevented with simple measures such as the adoption of a balanced diet, regular physical activity, and maintenance of the ideal weight [2]. However, the industrial way of life has been contributing for unhealthy lifestyles and obesity increasing, especially in urban and industrialized countries, which is intrinsically related to increasing breast cancer incidence rates in the next decades [1].

Although the amount of breast cancer cases in economically developed regions is increasing, mortality is decreasing due to the availability of early detection technologies, from self-examination campaigns to image-based diagnostic technologies, especially mammography [1].

In underdevelopment countries, however, the increasing incidence of breast cancer has been accompanied by the augment of the mortality rates. In East Africa, the incidence is 30 new cases per 100.00 women per year, whilst in Western Europe and industrialized, economically developed regions of the world, the incidence of breast cancer has reached more than 90 new cases for each group of 100.00 women per year [1]. The mortality rates, however, are almost identical in these two regions, about 15 per 100.00 women [1]. One of the causes is that patients from East Africa do not have easy access to image diagnosis. Therefore, breast cancer is usually detected in advanced stages. Additionally, this fact may lead to the need of mastectomies, mutilating surgeries in which the suspicious mammary tissue is completely removed [3]. Consequently, in order to promote early detection of breast cancer, self-examination by touch is not sufficient:

availability of imaging diagnosis technologies is fundamental once, in some cases, tumors take about 10 years to become palpable [3].

By the importance of breast imaging techniques, professional radiologist experience assumes a crucial role at finding and interpreting clinical data and developing accurate diagnosis. This is a particularly complex task due to the wide variability of cases, where many do not accurately fit in classical models and descriptions [4].

Particularly, there is a large difficulty about interpreting masses in noisy images such as those resulting from a mammography acquisition [5][6]. Furthermore, several studies indicate that over than 70% of breast cancer biopsy surgeries returns benign findings [7]-[9]. In order to turn diagnosis less susceptible to errors, several decision support systems have been developed to aid at the detection and classification of mammary lesions by analyzing mammographic images, once mammography is still the most effective clinical method for the early detection of breast cancer [3]. Decision support systems can be important allies of health professionals in order to perform accurate diagnostic decision making.

The task of detect and classify mammary lesions using mammograms is highly dependent on the feature extraction stage, in which the regions of interest, clinically determined by using specialist knowledge, are pre-processed, and moments, statistics, and other measures are extracted. In breast cancer applications the use of texture descriptors combined with segmentation methods is very common. However, more complex pre-processing approaches have being used in order to reach higher classification rates by modifying feature dimensionality. One of the most successful techniques is the series of wavelets: each image of region of interest is decomposed by a series of details images with different resolutions and a reduced and simplified version of the original image [10]. Features are extracted from these image components. Several works point this multi-resolution approach as able to detect mammary lesions and, in some cases, classify them as benign or malignant findings, with successful differentiation of the suspicious lesion and the mammary tissue [10].

Zernike moments have been widely used to perform mass shape analysis [12][13]. The lesion shape is essential for the determination of the malignancy degree of mammary lesions [15].

Herein this work we propose a methodology to detect and classify mammographic images by using multi-resolution wavelets decomposition and Zernike moments, without the needing of manual segmentation or adjustment, in the feature extractor stage, and kernel-based neural networks, namely Support Vector Machines (SVM) [33] and Extreme Learning Machines (ELM) [14], in the classification stage. We investigated the use of the more widespread wavelets and how they could affect detection and classification performance, particularly the classification rate and the training time. We varied neural networks kernels as well. We claim that the combination of multi-resolution wavelets and Zernike moments can mix texture and shape features, improving image representation for the classification stage. We also claim that the use of SVM and ELM furnish accurate classifiers just varying the kernels of the hidden layer neurons, in order to optimize classification accuracy.

For the experiments, we used the classification criteria for mammographic images defined according to the American College of Radiology, described on BIRADS (Breast Imaging Reporting and Data) [15]. The goal of BIRADS is to group the cases into three classes: normal (i.e. without cancer), benign and malignant lesion. We used 355 images of fatty breast tissue of IRMA database [17], with 233 normal instances, 72 benign, and 83 malignant cases. We generated synthetic instances of benign and malignant cases using linear combinations with random weights, in order to balance our database, getting a final amount of 699 instances, with 233 instances for each class. Classification was performed using SVM and ELM networks with modified kernels, in order to optimize accuracy rates, reaching 94.11%.

Learning is usually a time consuming affair as it may involve many iterations through the training data [16], involving cross validation combining with different random initial conditions, testing different modified kernels (learning functions), and different initial weights of layers neurons when neural networks are employed as learning technique. Also, whenever a new data is received, batch learning uses the past data concatenated with the new data and performs a retraining, thus consuming a lot of time [16]. Furthermore, long time is necessary in order to process the new image cases on the feature extraction stage. A solution in order to reduce the training time is the online learning which is only necessary to processing and training the new data (images) without retraining the past data. Online learning, however, shows lower accuracy results than batch learning, in all studied cases [16].

Our proposed model can combine high accuracy rate with low learning time in a batch learning approach. Considering both accuracy rates and training times, we defined the ration between average percentage accuracy and average training time in a reverse order. Our proposal was 50 times higher than the ratio obtained using the best method of the state-of-the-art.

The average training times are shown in seconds, in section 6. In fact, however, training time can consume days. The reason is that learning employs all parameters (wavelets family) exploration on image processing, cross-validation, different random initial conditions and weights, beyond testing different modified kernels. Then, whenever a new data is received, it can be necessary days in order to choose the best configuration, involving choosing the best feature extraction process, classifier, kernel, besides validation of methodological cares. As our proposed model can combine high accuracy rate with low learning time, whenever a new data is received, our work will be able to save a lot of time, hours, in learning process in relation to the best method of the state-of-the-art.

Several works have being developed to solve the problem of detect and classify mammary lesions using mammograms, focusing feature extraction. Table 1 presents a list of some important works of the state-of-the-art, organized taking into account the ability to detect and classify lesions, the need of human segmentation, a brief description of the feature extraction, and the amount of databases used in the experiments.

The works listed on Table 1 utilize different mammography databases. Selecting which database should be used is essential to get meaningful results, once getting satisfactory results using a determined database does not necessarily mean success when using another one. In this work we used IRMA (Image Retrieval in Medical Applications), a mammography database composed by two public databases, MiniMIAS (Mammographic Image Analysis Society database) and DDSM (Digital Database for Screening Mammography), and private mammograms [17]. Therefore, this avoids methodological questions on the achieved accuracy rates, in relation to the set of used images.

**Table 1 - Summary of main techniques for breast mass detection and/or classification.**

| *Authors* | *Description* | *Database Amount* | *Detection* | *Classification* | *Human Segmentation* |
|---|---|---|---|---|---|
| Proposed model | Morphological wavelets | 1* | Yes | Yes | No |
| Nascimento et al. (2013) [10] | Wavelets | 1 | Yes | Yes | No |
| Saki et al. (2011) [12] | Zernike moments | 1 | No | Yes | Yes |
| Saki et al. (2013) [13] | Spiculation index | 1 | No | Yes | Yes |
| Rouhi et al. (2015) [18] | CNN segmentation | 2 | No | Yes | Yes |
| Wang et al. (2014) [19] | Region growing | 1 | Yes | No | Yes |

Table 1 illustrates which methods of the state-of-the-art are able to perform detection, and which ones are designed to perform classification. Detection means the ability to differentiate between normal tissue and lesion. Classification means the capacity to differentiate between benign and malignant lesions. Wang et al. (2014) proposed a method for the detection of lesions based on region growing [19]. Saki et al. (2011), Saki et al. (2013), and Rouhi et al. (2015) proposed methods to classify lesions between benign and malignant lesions [12][13][18]. Saki et al. (2011) used Zernike moments as shape descriptors [12]. Saki et al. (2013) employed the spiculation index to represent the geometry of lesion boundary [12]. Rouhi et al. (2015) uses CNN segmentation at the feature extraction stage [12]. In all these methods human segmentation is needed [12][13][18] [19].

Table 1 shows the needing of human segmentation during training (learning phase) and/or using phases. Wang et al. [19] propose a methodology that requires human adjustment in order to determine the seed of region growing technique. The technique seeks similarities among the neighboring pixels of seed. The neighborhood of neighboring pixels of seed is also analyzed. Then, the initial grouping is expanded until no more pixels have to be added. Obviously, a maladjusted seed generates inadequate results. Therefore, human adjustment is present both in training and using phases. Though, human adjustment can make the process lengthy, stressfully and more error-prone. In addition, health professionals may have difficulties at handling a computational tool to extract features from medical images.

In other works, the radiologist is responsible to identify the mass edge [12][13][18]. After edge detection, there is the study of area occupied by the mass and its neighborhood. The expectation is that, after training phase, the computational model becomes able to detect the edge lesion automatically. If the edge automatic identification is not of good quality, obviously the mass classification will be compromised.

The works that require specialist segmentation do not deal with the original image. The mass of interest is detected and segmented from other mammography information directly by the radiologist. Once these images, handled by specialists, are not disclosed, the replica of this type of project becomes impractical. Their achieved results are not totally comparable to others, due to the use of distinct metrics. Therefore, applications highly dependent on human segmentation are not easily reproducible.

Nascimento et al. (2013) proposed a detection and classification method based on wavelets decomposition: the image of the region of interest given by database, i.e. the suspicious region, is decomposed in a series of wavelets and, afterwards, energy moments are extracted, being classified by a SVM network [10]. As such an approach avoids the need of specialist adjustment or segmentation, we adopted it and replicated it this work in order to improve this approach. Nascimento et al. (2013) compared three types of functions based on wavelets: Bi-orthogonal 3.7, Daubechies-8, and Symlet 8 [10]. The decomposition of the 128x128 mammogram is performed at two levels. Then the images from the first and second levels are adjusted to the dimensions 64x64 and 32x32, respectively. The strategy adopted by Nascimento et al. (2013) is to smooth the image components by calculating the local energies normalized by the total image energy. Afterwards, the SVD (Singular Value Decomposition) technique is applied over all the smoothed images, in order to reduce the amount of data necessary to form the feature vectors and, consequently, reducing the dimensionality of the classification problem.

This work is organized as following: in section 2, ,we present some of the fundamentals for classification and comments on the image database we used; in section 3, we explain the mammograms feature extractors employed; in section 4, we present the methodology we are proposing; in section 5, we make comments on the neural networks we used as classifiers; in section 6, we show the experimental results and some discussion; finally, in section 7, we make general conclusions and discuss perspectives of our work.

## 2. Materials and methods

*2.1. Breast and Mass Classification*

IRMA database description details the breast region description, as can be seen in Fig. 1. In young women breasts, there is usually low adipose (fat) tissue: the breasts are dense. Muscles and others breast tissues (parenchyma) occupy most part of the breast.

In a dense breast, parenchyma occupies more than 75% of breast tissue. In a predominantly dense breast, parenchyma occupies between 50% and 75% of breast tissue, whilst, in a predominantly adipose breast, parenchyma occupies between 25% and 50% of breast tissue. In an adipose breast, there is low parenchyma, less than 25% of breast tissue, and much more adipose tissue.

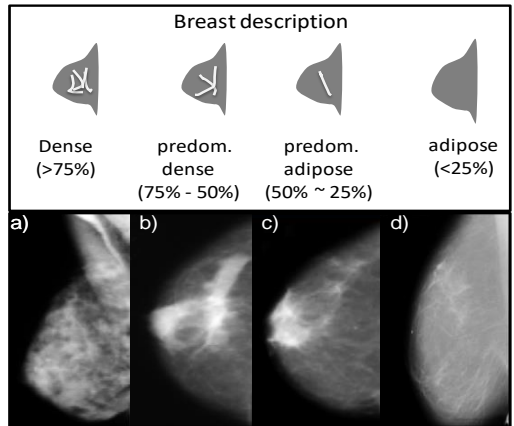

**Fig. 1. Breast classification and description.**

Adipose breasts are characteristic of elderly patients. There is a tendency of muscles and other parenchyma tissues being replaced by adipose tissue.

IRMA database description also provides mass classification information according to BIRADS criteria. Masses are classified into five groups: regular, lobular, microlobulated, irregular and spiculated, as can be seen in Fig. 2. A regular mass has benign characteristics, while a spiculated lesion has very high chances of being malignant. A regular mass presents a smooth edge with few variations. A lobular lesion has a wavy contour. A microlobulated mass presents smalls waves in the edge. A spiculated lesion has radiant lines in its margins. Finally, an irregular mass has, obviously, an irregular shape. An irregular mass does not fit the description of any other group.

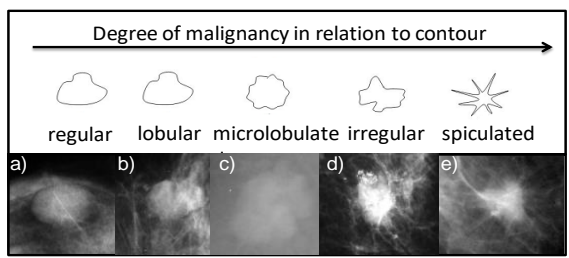

**Fig. 2. Classification of breast masses according to their contours.**

In relation to density, Brazil's National Institute of Cancer (INCA) classifies masses into five groups: heterogeneous, fat, low density, isodense and high-density, as shown in Fig. 3. A high-density mass presents a density higher than tissue density. Isodense masses have density equals to skin. Low-density masses have density lower than tissue density. Fat lesions have low-density and contain fats in their neighboring. Finally, heterogeneous masses have low-density and are partially occupied by mammary parenchyma.

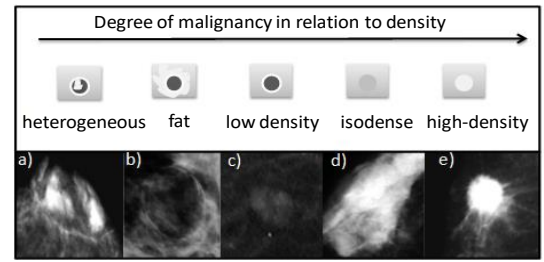

**Fig. 3. Classification of breast masses according to their densities.**

Theoretically, malignant tumors have a higher density than tissue density, whilst benign masses have lower density. BIRADS, for instance, mentions that fat lesions are massively benign [15].

### 2.2. Database

IRMA database is composed by some private mammography databases and two public databases, namely [17]:
- Mini-MIAS (Mammographic Image Analysis Society database) [20],
- DDSM (Digital Database for Screening Mammography) [21].

The database documentation details the scanning method of breast imaging, presenting the imaging directions. These directions can be mediolateral and cranio-caudal. IRMA documentation also specifies if it was imaged the right or the left breast. Lesions and breast tissues are also classified according to BIRADS criteria, as mentioned before [15].

All available database cases were investigated. There are not heterogeneous, fat and low-density masses, explained in the previous sections. All lesions are isodense and high-density.

It is quite common the presence of undesirable artifacts in digital mammograms due to the lack of care at the digitalization or unexpected variations inherent to the acquisition process, e.g. calibration problems. These artifacts are usually present as random elements over the breast, like tapes, trowels, and handwritten and digital notes. Furthermore, it is also common grooves and greases caused by fingerprints. These elements contribute to increase image analysis difficulties. IRMA database avoids these situations by furnishing the regions of interest (ROI) clinically selected using specialist knowledge.

### 2.3. Balancing Databases

IRMA also provides cases of calcification, architectural distortion, and asymmetry. We did not deal with these type of lesions. The area occupied by a calcification, for example, is very small in comparison with the breast area. Our proposal addresses only the adipose (fatty) breast tissues, because mammary parenchyma gray levels in dense breasts images are very close to mass regions gray levels. In dense breasts cases, usually of young patients, commonly it is not possible to identify the edge lesion, including for experienced radiologists.

Given the appropriate restriction, IRMA provides 355 cases of adipose (fatty) breasts. There are 233 normal cases (normal tissue, without lesions), 72 benign mass cases and 83 malignant lesion cases. In order to use IRMA database in classification applications, it is necessary to balance this dataset, generating 161 and 150 synthetic benign and malignant instances, respectively. The proposed methodology is creating artificial instances based on real image features instances. Therefore, a vector of random weights is created, with the same amount of artificial class cases. The weights uniformly vary from 0 to 1. These weights are used to form linear combinations of the real instances, in order to calculate the synthetic instances.

## 3. Image feature extraction

### 3.1. Morphological Spectrum

Mathematical Morphology is a complete theory of nonlinear processing extensively used in digital image processing. It is based on shape transformations preserving the object relations of inclusion. Its fundamental operators are called *erosion* and *dilation* [34]-[37]. Furthermore, Mathematical Morphology can be considered a constructive theory, because all operators are built using erosions and dilations as bases. It turns possible the construction of several operators designed to specific applications, as image filtering, feature extraction and others [34]-[37]. Mathematically, erosion and dilation are formalized in accordance with the expressions of Eq. 1 and 2, respectively [34]-[37].

$$\varepsilon_g(f)(u) = \bigcap_{v \in S} f(v) \vee \overline{g}(u-v), \quad (1)$$

$$\delta_g(f)(u) = \bigcup_{v \in S} f(v) \wedge g(u-v), \quad (2)$$

where $f: S \to [0,1]$ and $g: S \to [0,1]$ are normalized monospectral images with support $S \subseteq Z^2$, the operators $\cup$ and $\vee$ are associated with the maximum operator, whilst $\cap$ and $\wedge$ are associated with the minimum operator; $\overline{g}(u) = 1 - g(u)$ is the negation of $g(u), \forall u \in S; g: S \to [0,1]$, is the structure element of the respective erosion and dilation [34]-[37].

The erosion transforms the original image in such a way to make the negative of the structure element to incase in the original image, i.e. the original image $f(u)$ will be modified to make areas similar to $\overline{g}(u)$ increase, to incase $\overline{g}(u)$ [37]. If we associate the 1's to the absolute white and the 0's to the absolute black, we will perceive that the erosion results in the augment of the darkest areas and the elimination of the brightest areas [37].

The dilation transforms the original image in such a way to make the structure element to incase in the original image, i.e. the original image $f(u)$ will be modified to make areas similar to $g(u)$ increase, to incase $g(u)$ [37]. If we associate the 1's to the absolute white and the 0's to the absolute black, we will perceive that the dilation results in the augment of the brightest areas and the elimination of the darkest areas [37].

The implementation of a dilation of an image $f(u)$ followed by an erosion, with the same structure element $g(u)$, is known as *closing* $\phi_g(f)(u)$. The application of an erosion followed by an dilation, also with the same structuring element is known as *opening* $\gamma_g(f)(u)$ [37].

Let us consider a normalized monospectral image $f: S \to [0,1]$. The residual area resulting from the $k$-th opening operation, where $k \geq 0$, by the structure element $g: S \to [0,1]$, is given by the expression on Eq. 3 as following [37].

$$V(k) = \sum_{u \in S} \gamma_g^k(f)(u). \quad (3)$$

The mathematical expression in Eq. 4 describes in details the discrete accumulated density function, $\Xi: Z_+ \to [0,1]$ associated to image $f: S \to [0,1]$ [37].

$$\Xi[k] = 1 - \frac{V(k)}{V(0)}. \quad (4)$$

The discrete density function, also known as the *pattern spectrum* or *morphological spectrum*, is presented on Eq. 5 [37]. The pattern or morphological spectrum can be used as a very precise feature extractor based on shape, because the morphological theory guarantees that each binary image has unique representation based on this spectrum [37]. The pattern spectrum can be employed in pattern recognition applications as a type of digital signature [37].

$$\xi[k] = \Xi[k+1] - \Xi[k]. \quad (5)$$

In order to calculate the morphological spectrum, we have to perform the operations until the algorithm converges, i.e. when we get a null image, which returns null residual area ($V(k) = 0$). Different mammograms require $k$-different generations in order to conclude the morphological spectrum. Therefore, trying to avoid the problem of different-sized morphological spectra, we represented these spectra by using seven statistics, namely: mean, standard deviation, mode, median, kurtosis, minimum, and maximum value.

Let us consider a sphere of radius R, with a structure element in a disc shape. The correspondent morphological spectrum will be an impulse. By changing the radius size to 2R, the spectrum will remain to be an impulse. This fact illustrates the singular characteristic of morphological spectra as unique shape identifiers. Therefore, irregularities in mammary mass edges results in low amplitude variations in morphological spectra.

Morphological spectra may be useful in mammography study, because it present as advantage the ability to disregard low amplitude variations in mass edges. Furthermore, pattern spectra can be used as unique representations of tumor shape, regardless of size.

### 3.2. Image representation using wavelets

A wavelet $\psi(t)$ is a wave-like function which obeys to the following conditions:

$$\int_{-\infty}^{+\infty} \psi(t) dt = 0, \quad (6)$$

$$\int_{-\infty}^{+\infty} |\psi(t)|^2 dt < +\infty. \quad (7)$$

The wavelet $\psi(t)$ is called a mother wavelet, with child wavelets $\psi_{a,b}(t)$ translated and scaled versions of the mother wavelet, given by:

$$\psi_{a,b}(t) = \frac{1}{\sqrt{a}} \psi\left(\frac{t-b}{a}\right), \quad (8)$$

where $a$ and $b$ are the scaling and the translating factors, respectively. Given a continuous time signal $x(t)$, its continuous wavelet transform, $X(a,b)$, is given by:

$$X(a,b) = \int_{-\infty}^{+\infty} x(t) \psi_{a,b}^*(t) dt, \quad (9)$$

where * means the conjugate.

Wavelets are very powerful tools to represent images using a multi-resolution approach. The wavelet transform applied to the image processing can be implemented in a two-dimensional way. Mallat proposed a discrete wavelet transform of a unidimensional signal through the decomposition a series of signal components generated by discrete high-pass and low-pass filters [22].

A similar procedure can be applied to decompose images in order to get a multi-resolution representation. Fig. 4 shows the two-dimensional wavelet decomposition algorithm considering just one level of resolution. The image $A_j$ is convoluted by $h(n)$ and $g(n)$ filters. The two images, resulting from the convolution, are decimated by a factor of 2, where odd lines and columns are excluded. Afterwards, in the end of each level, Mallat's algorithm generates an approximation image, obtained by low pass filtering, and three detail images, resulting from high pass oriented filtering.

An approximation image (LL) consists of a representation of the original image without details, resulting from the application of a low pass filter. It appears to be a blurred version of the original image, due to the reduction of the number of gray levels.

A detail image aims to represent the original image in high frequency. A high pass filter enhances the elements present in the edge of the original image. At each level of wavelets, Mallat's algorithm generates three detail images: one to horizontal details (HL), another to vertical details (LH), and finally, to diagonal details (HH). The next level decomposition process uses, as source, the approximation image from the level immediately preceding.

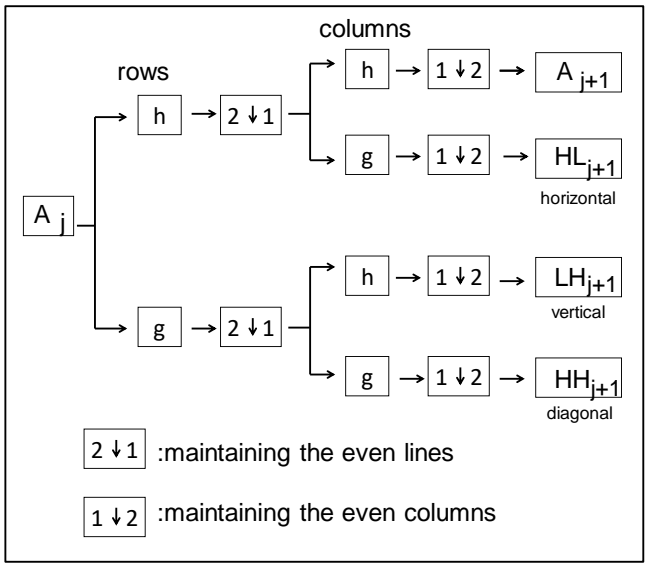

**Fig. 4. Wavelet decomposition algorithm for two-dimensional wavelets, considering just one resolution level.**

Fig. 5 presents the wavelets decomposition process into levels, using Daubechies 8. The goal of this technique is to decompose the original image in an approximation image and three detail images, at each level.

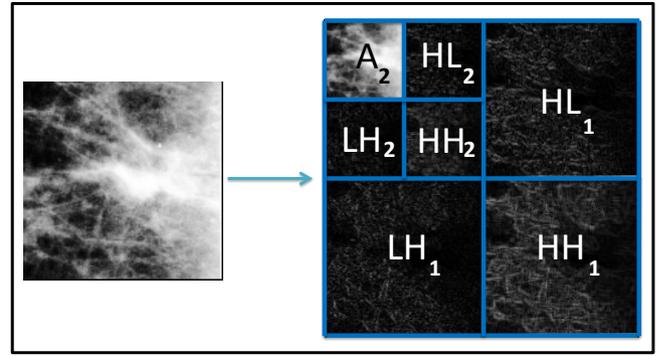

**Fig. 5. Illustration of wavelet decomposition in 2 levels, based on Daubechies 8 function.**

### 3.3. Zernike Moments

The Zernike polynomials are a basis of complex orthogonal polynomials [11]. The mathematical expression of the radial Zernike polynomials is the following, illustrated by Eq. 10:

$$R_{n,m} = \sum_{s=0}^{(n-|m|)/2} (-1)^s \frac{(n-s)!}{s!(((n+|m|)/2)-s)!(((n-|m|)/2)-s)!} p^{n-2s}, \quad (10)$$

where $n$ is a non-negative integer representing the radial polynomial order; $m$ is a non-zero integer, for $n - |m|$ non-negative and even; $p$ is the distance between the center and a point of the polynomial function. The computation of Zernike basis function, expressed in polar coordinates, is formalized by Eq. 11, as following:

$$V_{n,m}(p,\theta) = R_{n,m}(p)e^{jm\theta}, \lfloor p \rfloor \leq 1 \quad (11)$$

where $\theta$ is the angle formed between the line segment that joins the center to point p and the axis coordinate, also named the azimuthal angle. The imaginary unit is given by $j = \sqrt{-1}$.

Since Zernike polynomials integrate an orthogonal basis, Zernike moments are able to represent image properties with no redundancy or overlap of information between the moments. Zernike moments are significantly dependent on scaling and translation. Nevertheless, their magnitudes are independent of the rotation angle of the object. Therefore, we can utilize them to describe shape characteristics of the main objects. Due to these qualities, Zernike moments have being used successfully to represent mammary regions of interest in mammograms [12][13]. Fig. 6 illustrates the magnitude of the first Zernike moments in the unit disk.

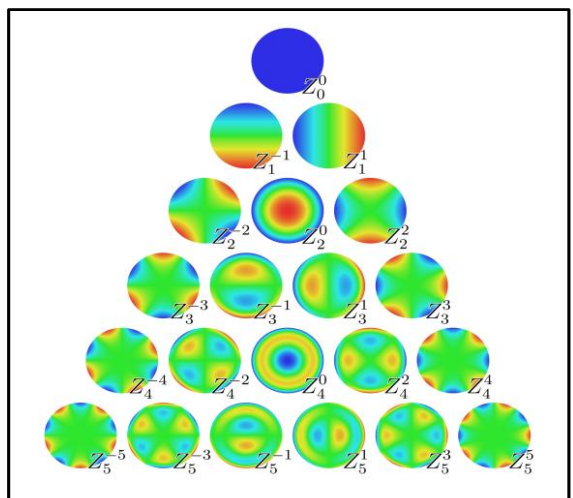

**Fig. 6. Plots of the magnitude of Zernike basis functions in the unit disk. Figure extracted from [12].**

Zernike moments are described in Table 2. The order $n$ ranges from 3 to 10, in ascending order. The iteration module $m$ should be less or equal to the given order $n$. All the differences $n-|m|$ must be multiple of 2.

**Table 2 - The Zernike Moments studied.**

| *Zernike Moments* | | |
|---|---|---|
| *Order(n)* | *Iteration (m)* | *Number of moments* |
| 3 | 1,3 | |
| 4 | 0,2,4 | |
| 5 | 1,3,5 | |
| 6 | 0,2,4,6 | 32 |
| 7 | 1,3,5,7 | |
| 8 | 0,2,4,6,8 | |
| 9 | 1,3,5,7,9 | |
| 10 | 0,2,4,6,8,10 | |

Therefore, we can utilize them to describe shape characteristics of the objects of interest. Due to these qualities, Zernike moments have being used successfully to represent mammary regions of interest in mammograms [12][13]. Note that Zernike moment, with $n=0$, $m=0$, has an adequate shape in order to describe regular masses, while the moment with $n=5$, $m=5$, can describe spiculated mass which has lines from its center in direction to its margins, seen in Fig. 2. Fig. 7 shows an example of mass shape being descript by a Zernike moment, with $n=0$, $m=0$.

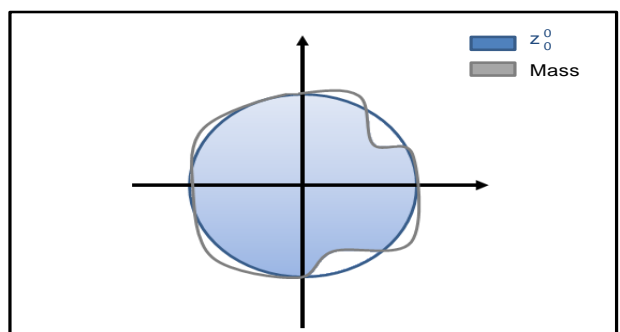

Fig. 7. Description of mass shape and Zernike, *n=0; m=0.*

## 4. Methodology

Fig. 8 shows the block diagram of the proposed design flow. First of all, mammographic images are preprocessed and have their histogram normalized, in order to associate the brightest pixels to absolute white and the darkest pixels to absolute black. Afterwards, images are decomposed onto series of wavelets, according to Mallat's algorithm. Herein this paper we used a wavelet architecture composed by four levels, from which we obtained 13 image components. Hence, we extract 32 Zernike moments from each wavelet image component. These 32 descriptors, for each image component, are then normalized between 0 and 1.

The discrete low-pass filter causes a blurring in Approximation images $A_{level}$, with larger intensity as level decomposition increases. As there is advance of levels, Approximation images $A_{level}$ lose resolution. Then, in first level, there is study of masses of all dimensions. While at higher levels, there is evaluation of large lesions since the small shapes are being blurred and attached to background. Then, our methodology works in a multi-resolution approach. Herein this work we create a wavelet's architecture composed by four levels, from which we obtained 13 image components. We claim that this strategy is able to deal with several mass sizes.

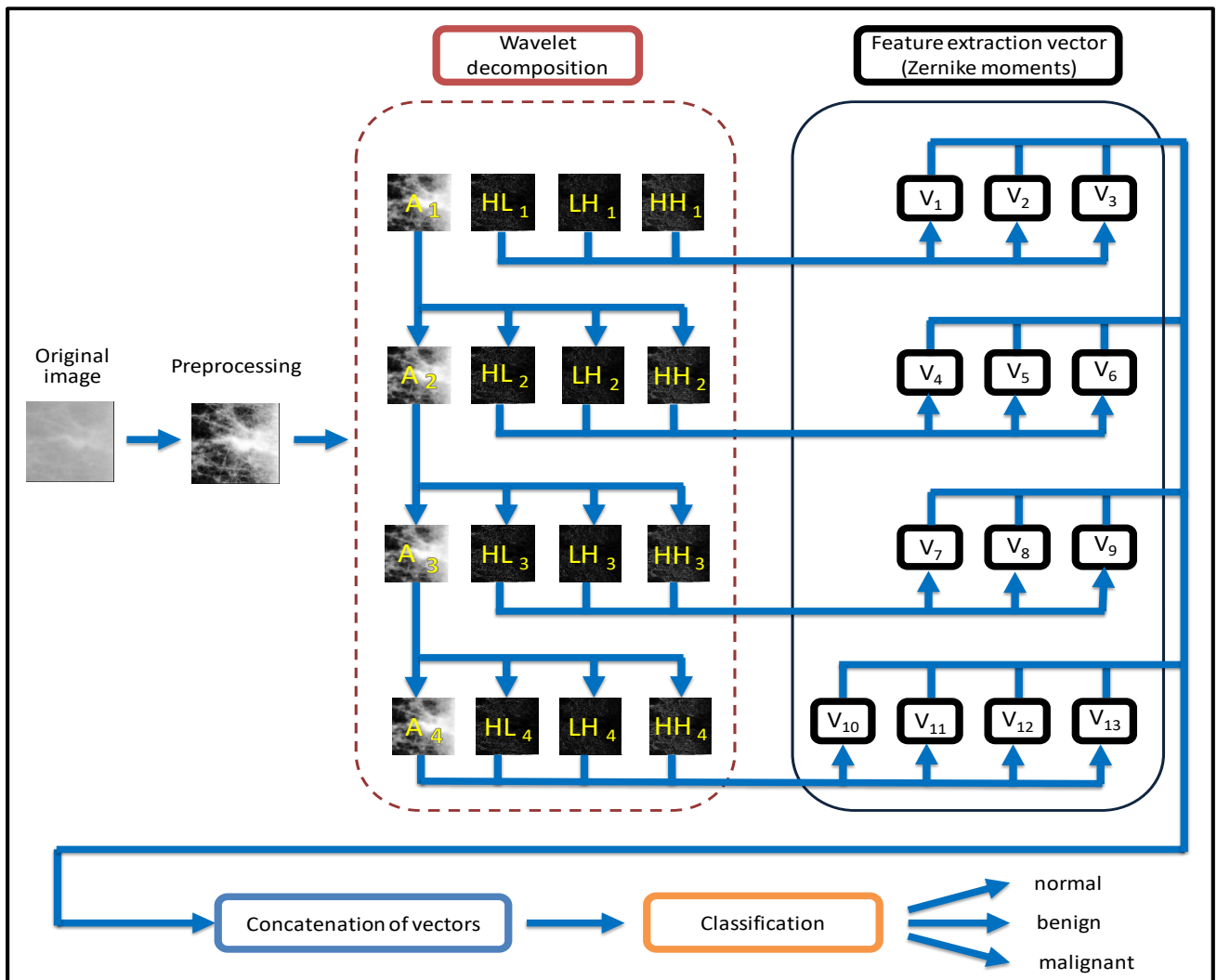

**Fig. 8. Block diagram of the proposed breast lesion detection and classification methodology.**

The creation of our feature extraction stage is not based on empiricism, or randomly selected in trial and error method. There was a deeper study of *American College of Radiology* rules, explained in section 2. Wavelets investigate image texture intensity, in multi-resolution. It is noteworthy that texture is essential in order to determine the lesion presence or absence, with several sizes. Series of wavelets is able to describe texture accurately

Taking into account mass textures (densities) and classical descriptions, there is usually an abrupt distinction in the intensity of gray levels between benign and malignant lesions, showed in Fig. 3. However, on mammographic images provided by IRMA database, it hardly ever occurs, because all masses have a tendency to be isodense or even high-density ones. Consequently, textures use to be very close regarding gray levels. Then, wavelets decomposition, in isolation, may be unable to detect and classify masses.

It increases the importance of Zernike moments, in order to improve the capabilities of studying malignancy degrees of breast cancer. These aspects contribute to our proposal being able to both detect and classify breast masses between benign and malignant findings. Zernike moments studies mass shape. Zernike moments are projected in images from wavelet decomposition. The classification between malignant and benign lesions uses as parameter the results of these projections. Zernike moments could give a non-intuitive set of shape descriptors, described in Fig. 2. Thus, series of wavelets are able to describe texture (density) in a multi-resolution approach, and Zernike is important in order to describe mass shape (contour). Our methodology has the interesting feature of incorporating texture as well as shape.

## 5. Classifiers

The proposed breast cancer detection and classification methodology employs artificial neural networks as classifiers. We used the following network architectures: Multilayer Perceptrons (MLP), Extreme Learning Machines (ELM), and Support Vector Machines (SVM). ELM and SVM networks were adopted using a multi-kernel approach, i.e. the kernels of the hidden layer neurons were changed in order to discover the best configuration for the proposed feature extraction method, regarding classification performance, especially accuracy. These architectures were also chosen to allow comparisons with the state-of-the-art methods.

For all neural networks architectures, we set 416 inputs, i.e. the dimensionality of the feature vectors: 416 features, with 32 Zernike moments for each of the 13 wavelet image components. The number of neurons in the hidden layer was set according to each type of network. For all neural networks we used, the number of neurons of the output layer equals the number of classes: three neurons, each one dedicated to identify one of the classes of interest established by BIRADS criteria: normal (no lesion), benign, and malignant lesion. Furthermore, we use Fit Decision Tree [23] and Fit *k*-nearest Neighbor Classifiers [24]. They are popular classifiers used in state-of-the-art [10].

### 5.1. Neural Networks

#### 5.1.1. Multilayer Perceptrons

This proposal employs 11 different learning functions based on MLP classifiers. In MLPs networks with backpropagation, there is a concern about how to prevent the effects of local minima. Some approaches consist of adding procedures to control weights adjustment to escape from these regions. These procedures usually increase the training time necessary to reach high classification accuracies.

In order to train and test MLPs, we separated 70% of data for training, 15% for cross-validation, and 15% for absolute testing. The maximum number of epochs for training is 100. Cross-validation is run after each season. There is a tolerance of 5 failures for each cross-validation run. The total training time includes, obviously, the time spent in effective training and the cross-validation time.

We employed three types of architectures for each MLP network. The first architecture uses a single hidden layer with 100 neurons. The second architecture uses a single hidden layer with 500 neurons. The second architecture results are symbolized by an asterisk in tables' results. Finally, the third architecture uses two hidden layers, each one with 100 neurons. The third architecture results are symbolized by a double asterisk. As mentioned before, we analyzed 11 different learning functions and metrics of neural networks based on backpropagation, as following:

1. Batch training with weight and bias learning rules: trains a network with weight and bias learning rules with batch updates. The weights and biases are updated at the end of an entire pass through the training data [25].
2. Conjugate gradient backpropagation with Powell-Beale restarts: is a network training function that updates weight and bias values according to the conjugate gradient backpropagation with Powell-Beale restarts [26].
3. Conjugate gradient backpropagation with Fletcher-Reeves updates: employs a training function that updates weight and bias values according to conjugate gradient backpropagation with Fletcher-Reeves updates [27].
4. Conjugate gradient backpropagation with Polak-Ribiére updates: applies network training function that updates weight and bias values according to conjugate gradient backpropagation with Polak-Ribiére updates [28].
5. Gradient descent backpropagation: is a network training function that updates weight and bias values according to gradient descent [25].
6. Gradient descent with momentum backpropagation: employs a network training function that updates weight and bias values using the gradient descent with momentum. It allows a network to respond not only to the local gradient, but also to recent trends in the error surface [25].
7. Gradient descent with adaptive learning rate backpropagation: applies a network training function that updates weight and bias values according to gradient descent with adaptive learning rate. The performance of the algorithm is very sensitive to the proper setting of the learning rate. If the learning rate is set too high, the algorithm can oscillate and become unstable. If the learning rate is too small, the algorithm takes too long to converge. The optimal learning rate changes during the training process [25].
8. Gradient descent with momentum and adaptive learning rate backpropagation combined: is a network training function that updates weight and bias values according to gradient descent momentum and an adaptive learning rate. It combines adaptive learning rate with momentum training [25].
9. One-step secant backpropagation: uses a network training function that updates weight and bias values according to the one-step secant method [29].
10. Resilient backpropagation (Rprop): applies a network training function that updates weight and bias values using the resilient backpropagation algorithm (Rprop) [30].
11. Scaled conjugate gradient backpropagation: is a network training function that updates weight and bias values according to the scaled conjugate gradient method [31].

For each learning function we employ 3 types of architectures, as stated earlier. Furthermore, we explore 30 different random permutations of data order, with the seed of random generator varying from 1 to 30 in an incremental way. In each one of permutation of data order, we explore 30 different initial weights of layers neurons, with the seed of random generator varying from 1 to 30 in an incremental way.

Then, for each learning function there are 2700 iterations. It involves 3 types of architectures, 30 different order of data for each one of architecture, and 30 different initial weights for each order of data.

#### 5.1.2. Extreme Learning Machines

ELM networks have as the main characteristic high training speed and good data prediction abilities when compared to other networks used as classifiers. ELM networks have a single non-recursive hidden layer. They have as a remarkable characteristic the fast training process, composed by just a few steps. The weights of the hidden layer neurons are randomly determined, once their kernels are defined. The weights of the output neurons are determined by the calculation of the generalized pseudoinverse of Moore-Penrose [32]. This learning process is performed in batch mode, where all data is presented to the network before the calculation of the weights, in a single iteration. Once this learning algorithm is not based on gradient descent methods, ELM networks are not affected by local minima. Furthermore, it is not necessary to define neither a learning rate parameter nor a maximum of iterations, once this is a non-iterative algorithm.

### 5.1.3. Support Vector Machines

SVM networks are single or multilayer networks with linear neurons in the output layer. These networks classify data by choosing the optimal hyperplanes in the output layer, which separate data within their respective classes, optimizing classification accuracy for the training set [33]. The best hyperplanes have the largest separation in relation to the given classes [33].

The original SVM training algorithm is able to deal with just two classes. Due to this limited binary nature, we used the LibSVM library, a computational library to classify data into multiple classes, once our problem of detection and classification of mammary lesions have three classes [33].

### 5.1.4. Multi-kernel configurations

Considering artificial neurons as basic connective processing units, kernels are mathematical functions which define how the output is calculated from the inputs and their synaptic weights. Linear kernels, as those used in McCulloch-Pitts original artificial neurons, define the output as a linear combination between the weights vector and the input vector. These neurons are not able to solve nonlinearly separable problems, e.g. separating Gaussian-distributed data. However, Radial Basis Function (RBF) kernels deal with these problems adequately, once neural outputs are calculated by applying the Euclidian distance between inputs and weights to a Gaussian function.

Herein our proposal we suggest the variation of the nature of hidden layer kernels in SVM and ELM networks, as an interesting strategy to maximize classification accuracy and minimize training time.

### 5.1.4.1. ELM configurations

The proposed work extends ELM networks to kernel learning. We tested 8 different types of kernels: RBF, linear, polynomial, wavelet transform, sigmoid, sine, hard limit, and tribas (triangular basis function) [32].

The proposed work has a methodological care on ELM configurations. We implemented cross-validation using the k-fold method with 10 folds. Our objective was avoiding the influence of training and test set on results. The dataset was divided into ten subsets. In the first iteration, the first subset is used for testing, while the others are reserved to training. This continues until all subsets are used for testing. We defined ELM total accuracy as the simple mean of the accuracies obtained in each k-fold step. The ELM architecture has 100 neurons on hidden layer.

For each kernel, we explore 30 different random permutations of data order, with the seed of random generator varying from 1 to 30 in an incremental way. In each one of permutation of data order, we explore 30 different initial weights of the layers neurons, with the seed of random generator varying from 1 to 30 in an incremental way.

Then, for each kernel there are 9000 iterations. It involves 30 different order of data, 30 different initial weights for each order of data, and 10 folds (iterations) for cross-validation.

### 5.1.4.2. SVM Configurations

The proposed work extends the SVM to learning based on kernels as well. The strategy employed here and the strategy we used in ELM networks were the same, considering architecture, training and test stages. We tested four different types of kernels: linear, polynomial, RBF, and sigmoid. We implemented cross-validation using the k-fold method with ten folds.

For each kernel, we explore 30 different random permutations of data order, with the seed of random generator varying from 1 to 30 in an incremental way. Then, for each kernel there are 300 iterations. It involves 30 different order of data, and 10 folds (iterations) for cross-validation.

### 5.2. Classifiers employed by state-of-the-art

### 5.2.1. Fit Decision Tree Configurations

The strategy employed by Fit Decision Tree is to find the best split on a categorical predictor for $k \geq 3$ classes. We implemented Decision Tree with $n$-1 for maximum number of splits. $n$ is the training sample size. We have a final amount of 699 instances (samples). We implemented cross-validation using the k-fold method with 10 folds. Then, the dataset is divided into ten subsets. From 1º to 9º subset, each one has 69 samples, the 10º subset has 78 samples. Then, from 1º to 9º $i$-iteration, the testing has 69 samples and training sample size is $n$=630. On 10 º iteration, the testing has 78 instances and training $n = 621$.

The minimum leaf size is 1. Fit Decision Tree splits all nodes in the current layer, and then counts the number of branch nodes. A layer is the set of nodes that are equidistant from the root node. If the number of branch nodes exceeds maximum number of splits, the algorithm follows its execution. The proposed split causes the number of observations in at least one leaf node to be fewer than minimum leaf size. We implemented cross-validation using the k-fold method with ten folds.

Fit Decision Tree employs 30 different random permutations of data order, with the seed of random generator varying from 1 to 30 in an incremental way. Then, we explore 300 iterations. It involves 30 different order of data, and 10 folds (iterations) for cross-validation.

### 5.2.2. Fit k-Nearest Neighbor Configurations

The training employed by Fit *k*-nearest Neighbor divide the training sample data into nodes with a maximum of 50 point per node. The dimensional space of each node is done in function of *f*-dimensional distance among points in a training dataset. We set 416 inputs for each instance (sample), i.e. the dimensionality among points of the feature training vectors *f* is obtained by calculated the Euclidian distance among features of each training sample.

After training, *k*-nearest neighbor classifier predicts a new instance in three steps. First, it is found the nearest training sample in relation to new instance. The "nearest score" concept is related to Euclidian distance among new point features and features of training instances. In second step, the nearest training sample class is investigated. Then, in third step, the new instance class assumes the nearest training sample class. We implemented cross-validation using the k-fold method with 10 folds. Fit *k*-nearest

Neighbor explore 300 iterations. Logic of calculation is the same as Fit Decision Tree.

## 6. Results

Herein this work we decided to replicate the proposal of Nascimento et al. (2013) [10] in order to perform comparisons between our proposal and such state-of-the-art method. We used three wavelets families, as proposed by Nascimento et al. (2013): Biorthogonal 3.7, Daubechies 8, and Symlet 8 [10].

We also investigated the performance of the morphological spectrum as part of the feature extraction process to describe breast lesions. We tested two 3x3 structure elements, square and cross, trying to investigate the effect of analysis given eight and four directions, respectively. Afterwards, seven statistics are measured from each pattern spectrum, in order to describe it, as explained in section 3.1.

Table 3 shows the results for average training and testing time, and average accuracy for all approaches in comparison with ours. The best cases are emphasized in bold. The first and second criteria are the sample mean average and the standard deviation, respectively.

Average training time is important in order to estimate the time to be spent whenever a new data is received. Learning (training) process can consume days in order to choose the best configuration, involving choosing the best feature extraction process, classifier, kernel, besides validation of methodological cares. Average testing time, on the other hand, is important in order to estimate the software response time.

SVM with linear, polynomial, RBF kernels are employed as classifier. Furthermore, we use Fit Decision Tree [23] and Fit $k$-nearest Neighbor Classifiers [24]. They are popular classifiers used in state-of-the-art [10].

From Table 3, considering the configurations of SVM with linear kernel as classifier and wavelets family Symlet 8 as part of our feature extraction, our proposal reached an accuracy of 94.11%. With the approach proposed by Nascimento et al. (2013), an inferior classification performance was reached, get an accuracy of 90.12% with wavelets family Daubechies 8. Furthermore, training times spent using the approach of Nascimento et al. (2013) were considerably higher than times obtained with other approaches.

Taking into account SVM with polynomial kernel, our proposal still take advantages over the works of the state-of-the-art. Using the wavelet function Symlet 8 both our proposal and the method of Nascimento et al. (2013) were able to achieve their best accuracies. In this case, our model achieved a performance of 91.51%, and Nascimento et al. (2013) reached an accuracy of 85.10%. In relation to training time, Nascimento et al. (2013) spent more times than our model.

Considering SVM with RBF kernel, accuracy rates are not feasible results independently of feature extraction technique. The morphological spectrum, with structure elements in a square format, achieved the better performance with rate of 50.04%. In other approaches, the RBF classifier misses more than hits, with performance slower than 50%. Taking into account training time, our model and Nascimento et al. (2013) reached very close results, with average training time of 0.29 seconds and 0.23 seconds, respectively.

In relation to Decision Tree, Nascimento et al. (2013) reached higher results than ours. Using the wavelet function Daubechies 8 both our proposal and the method of Nascimento et al. (2013) were able to achieve their best accuracies. Nascimento et al. (2013) reached an accuracy of 73.42% using the wavelet family Symlet 8. Our model achieved a performance of 69.56% with wavelet family Daubechies 8.

Taking into account K-NN classifier, our proposal reached an accuracy of 71.90% with wavelet family Biorthogonal 3.7. With approach proposed by Nascimento et al. (2013), a slightly higher classification performance was reached, getting a performance of 72.61% with also wavelet family Biorthogonal 3.7. In all evaluated feature extraction techniques, they are obtained suitable computational costs. Training times were about 0.01 seconds.

SVM with linear and polynomial kernels achieve best results both our work and Nascimento et al. (2013). Then, Table 4 and Table 5 show classes confusion for all feature extraction approaches using SVM linear and polynomial kernels, respectively. For BI-RADS, the class confusion is important to be distinguished. N., B., and M. are Normal, Benign, and Malignant abbreviations, respectively. Bio. 3.7, Db. 8, and Sym. 8 are Biorthogonal 3.7, Daubechies 8, and Symlet 8, abbreviations, respectively.

Booth Table 4 and Table 5 predicted classes are labeled on horizontal, while obtained classes are labeled on vertical. On Table 4, for example, on test dataset our proposed model, using wavelet family Biorthogonal 3.7 with SVM linear kernel, classified mistakenly, on average, 1.92 as normal cases but they are, de fact, benign masses.

On confusion matrix, the main diagonal is occupied by cases whenever obtained class coincides with expected class, named true positives cases. Then, a good classifier has main diagonal occupied by high values and others elements have low values. Table 4 and Table 5 show main diagonals emphasized in bold.

Table 6 shows Student's $t$-test between our proposal, using Symlet 8 wavelets, and the state-of-the-art approaches for test rate. The *Hypothesis* test is 1 for all comparison cases. Then, the null hypothesis is rejected at the 5% significance level. The two sets of data, in each experiment, are significantly different from each other. $p$-value of the test, return as a scalar value in the range [0,1]. $p$ is the probability of observing a test statistic as extreme as, or more extreme than, the observed value under the null hypothesis.

Fig. 9 shows boxplots related to the accuracy of the evaluated approaches. Regarding the SVM with linear kernel, we could perceive all techniques were able to reach reasonable results, except for feature extraction using morphological spectra.

Taking into account SVM with polynomial kernels our proposal could reach better results than the method of Nascimento et al. (2013), once we achieved higher accuracy rates. Considering SVM with polynomial kernel as well, feature extraction with morphological spectra was not a successful approach. In relation to SVM with RBF kernel, all feature extraction techniques are no able to reach reasonable results.

Even regarding to Fig. 9, our model and Nascimento et al. (2013) achieve close results when Decision tree and K-NN classifiers are employed. Feature extraction with morphological spectra again show lower results in comparison to other works.

Fig. 10 shows boxplots of training times obtained by the approaches we evaluated. Regarding SVM with linear kernels, the work of Nascimento et al. (2013) could not achieve reasonable results, once its average training time was considerably high. Furthermore, results obtained by using the approach of Nascimento et al. (2013) presented high dispersion.

Nevertheless, in the best cases, the learning process consumes much more time compared to other approaches.

In relation to polynomial kernels, our proposal had lower average training time, when compared to the work of Nascimento et al. (2013). Despite our proposal has presented large dispersion as well. Considering SVM with RBF kernel, Decision tree, and K-NN, training times were sensibly low for all approaches, with very low dispersions. Hence, in these cases, learning times were stable, without abrupt changes, independently of the data set arrangement. In all evaluated kernels, the morphological spectrum obtained suitable training times.

**Table 3. Comparison between the proposed model and the state-of-the-art approaches.**

| *Classifier* | *Feature extraction* | *Function* | *Train rate (%)* | *Test rate (%)* | *Train time (sec.)* | *Test time (sec.)* |
|---|---|---|---|---|---|---|
| SVM, linear | Proposed model | Biorthogonal 3.7 | 99.80 ± 0.18 | 91.82 ± 3.58 | 0.20 ± 0.06 | 0.02 ± 0.01 |
| | | Daubechies 8 | 99.80 ± 0.19 | 92.83 ± 3.54 | 0.18 ± 0.15 | 0.01 ± 0.01 |
| | | Symlet 8 | 99.80 ± 0.18 | **94.11 ± 3.00** | 0.18 ± 0.05 | 0.01 ± 0.01 |
| | spectrum (1) | | 54.31 ± 1.28 | 51.25 ± 4.90 | 0.02 ± 0.01 | 0.00 ± 0.01 |
| | spectrum (2) | | 51.49 ± 1.22 | 49.08 ± 5.60 | 0.03 ± 0.02 | 0.00 ± 0.01 |
| | Nascimento et al. [10] | Biorthogonal 3.7 | 99.33 ± 1.24 | 89.49 ± 4.23 | 7.43 ± 1.64 | 0.01 ± 0.01 |
| | | Daubechies 8 | 98.67 ± 1.88 | 90.12 ± 4.41 | 8.61 ± 0.61 | 0.01 ± 0.02 |
| | | Symlet 8 | 97.62 ± 1.91 | 89.18 ± 4.48 | 8.81 ± 0.60 | 0.01 ± 0.02 |
| SVM, polynomial | Proposed model | Biorthogonal 3.7 | 99.08 ± 0.38 | 89.79 ± 3.64 | 1.09 ± 1.61 | 0.01 ± 0.01 |
| | | Daubechies 8 | 99.04 ± 0.40 | 90.62 ± 3.69 | 0.68 ± 0.98 | 0.01 ± 0.01 |
| | | Symlet 8 | 99.04 ± 0.49 | 91.51 ± 3.43 | 0.92 ± 1.44 | 0.01 ± 0.01 |
| | spectrum (1) | | 56.45 ± 1.33 | 52.75 ± 4.73 | 1.83 ± 0.54 | 0.00 ± 0.01 |
| | spectrum (2) | | 55.49 ± 1.69 | 54.04 ± 5.87 | 2.59 ± 2.89 | 0.00 ± 0.01 |
| | Nascimento et al. [10] | Biorthogonal 3.7 | 92.94 ± 3.34 | 84.56 ± 5.08 | 10.19 ± 1.35 | 0.01 ± 0.02 |
| | | Daubechies 8 | 91.64 ± 3.44 | 84.10 ± 5.30 | 10.63 ± 1.22 | 0.01 ± 0.02 |
| | | Symlet 8 | 91.82 ± 3.24 | 85.10 ± 4.82 | 10.45 ± 1.19 | 0.01 ± 0.02 |
| SVM, RBF | Proposed model | Biorthogonal 3.7 | 98.92 ± 0.35 | 43.94 ± 6.46 | 0.29 ± 0.01 | 0.04 ± 0.01 |
| | | Daubechies 8 | 98.79 ± 0.39 | 44.63 ± 6.21 | 0.29 ± 0.01 | 0.04 ± 0.01 |
| | | Symlet 8 | 98.81 ± 0.38 | 44.63 ± 6.49 | 0.29 ± 0.01 | 0.04 ± 0.01 |
| | spectrum (1) | | 53.94 ± 1.21 | 50.04 ± 6.04 | 0.02 ± 0.01 | 0.00 ± 0.01 |
| | spectrum (2) | | 52.61 ± 1.22 | 48.89 ± 5.79 | 0.02 ± 0.00 | 0.00 ± 0.01 |
| | Nascimento et al. [10] | Biorthogonal 3.7 | 99.95 ± 0.08 | 28.27 ± 3.49 | 0.23 ± 0.01 | 0.03 ± 0.01 |
| | | Daubechies 8 | 99.96 ± 0.07 | 28.16 ± 3.45 | 0.23 ± 0.01 | 0.03 ± 0.01 |
| | | Symlet 8 | 99.96 ± 0.09 | 28.21 ± 3.51 | 0.23 ± 0.01 | 0.03 ± 0.01 |
| Decision tree | Proposed model | Biorthogonal 3.7 | 96.80 ± 0.68 | 64.86 ± 5.38 | 0.64 ± 0.08 | **0.00 ± 0.00** |
| | | Daubechies 8 | 96.94 ± 0.74 | 69.56 ± 5.50 | 0.59 ± 0.08 | **0.00 ± 0.00** |
| | | Symlet 8 | 97.13 ± 0.66 | 69.26 ± 5.39 | 0.63 ± 0.06 | **0.00 ± 0.00** |
| | spectrum (1) | | 87.86 ± 0.99 | 55.17 ± 5.88 | 0.03 ± 0.02 | **0.00 ± 0.00** |
| | spectrum (2) | | 89.87 ± 0.94 | 60.62 ± 5.60 | 0.03 ± 0.02 | **0.00 ± 0.00** |
| | Nascimento et al. [10] | Biorthogonal 3.7 | 95.94 ± 0.74 | 70.94 ± 5.18 | 0.45 ± 0.06 | **0.00 ± 0.00** |
| | | Daubechies 8 | 96.34 ± 0.79 | 71.96 ± 5.57 | 0.39 ± 0.03 | **0.00 ± 0.00** |
| | | Symlet 8 | 96.23 ± 0.77 | 73.42 ± 5.27 | 0.43 ± 0.06 | **0.00 ± 0.00** |
| K-NN | Proposed model | Biorthogonal 3.7 | **100.00 ± 0.00** | 71.90 ± 5.10 | **0.01 ± 0.01** | 0.02 ± 0.01 |
| | | Daubechies 8 | **100.00 ± 0.00** | 71.20 ± 5.15 | **0.01 ± 0.01** | 0.02 ± 0.01 |
| | | Symlet 8 | **100.00 ± 0.00** | 68.95 ± 5.29 | **0.01 ± 0.01** | 0.02 ± 0.01 |
| | spectrum (1) | | **100.00 ± 0.00** | 48.72 ± 6.06 | **0.01 ± 0.01** | 0.00 ± 0.01 |
| | spectrum (2) | | **100.00 ± 0.00** | 48.42 ± 6.07 | **0.01 ± 0.01** | 0.00 ± 0.01 |
| | Nascimento et al. [10] | Biorthogonal 3.7 | **100.00 ± 0.00** | 72.61 ± 5.18 | **0.01 ± 0.01** | 0.01 ± 0.00 |
| | | Daubechies 8 | **100.00 ± 0.00** | 70.66 ± 5.25 | **0.01 ± 0.01** | 0.01 ± 0.00 |
| | | Symlet 8 | **100.00 ± 0.00** | 70.20 ± 5.26 | **0.01 ± 0.01** | 0.01 ± 0.00 |

**Table 4. Confusion matrix of proposed model and state-of-the-art approaches as feature extraction and SVM neural network with linear kernel as classifier. N., B., and M. are Normal, Benign, and Malignant abbreviations, respectively.**

| *Feature extraction* | *Function* | | *Train* | | | *Test* | | |
|---|---|---|---|---|---|---|---|---|
| | | | N. | B. | M. | N. | B. | M. |
| | | N. | **209.70 ± 4.08** | 0.00 ± 0.00 | 0.00 ± 0.00 | **21.00 ± 3.87** | 1.92 ± 1.35 | 0.38 ± 0.73 |

| Feature extraction | Function | | Train | | | Test | | |
|---|---|---|---|---|---|---|---|---|
| | | | N. | B. | M. | N. | B. | M. |
| Proposed model | Bio. 3.7 | B. | 0.00 ± 0.00 | **209.20 ± 4.10** | 0.50 ± 0.92 | 1.11 ± 1.16 | **21.56 ± 3.83** | 0.63 ± 1.17 |
| | | M. | 0.00 ± 0.00 | 0.76 ± 1.12 | **208.94 ± 4.03** | 0.62 ± 0.83 | 1.05 ± 1.45 | **21.63 ± 3.73** |
| | Db. 8 | N. | **209.70 ± 4.08** | 0.00 ± 0.00 | 0.00 ± 0.00 | **21.37 ± 3.94** | 1.24 ± 1.15 | 0.69 ± 0.86 |
| | | B. | 0.00 ± 0.00 | **209.27 ± 4.07** | 0.43 ± 0.91 | 0.95 ± 1.06 | **21.82 ± 3.80** | 0.53 ± 1.09 |
| | | M. | 0.00 ± 0.00 | 0.83 ± 1.15 | **208.87 ± 4.08** | 0.52 ± 0.78 | 1.08 ± 1.42 | **21.70 ± 3.77** |
| | Sym. 8 | N. | **209.70 ± 4.08** | 0.00 ± 0.00 | 0.00 ± 0.00 | **22.14 ± 3.99** | 0.73 ± 0.81 | 0.43 ± 0.66 |
| | | B. | 0.00 ± 0.00 | **209.17 ± 4.09** | 0.53 ± 0.97 | 0.86 ± 0.99 | **21.87 ± 3.72** | 0.57 ± 1.06 |
| | | M. | 0.00 ± 0.00 | 0.71 ± 1.05 | **208.99 ± 4.06** | 0.50 ± 0.75 | 1.03 ± 1.40 | **21.77 ± 3.69** |
| spectrum (1) | | | N. | **183.33 ± 5.94** | 17.61 ± 12.63 | 8.76 ± 10.82 | **20.35 ± 3.72** | 1.97 ± 1.88 | 0.99 ± 1.63 |
| | | B. | 56.11 ± 4.84 | **93.63 ± 75.39** | 59.96 ± 73.29 | 6.24 ± 2.40 | **9.36 ± 7.78** | 7.70 ± 9.59 |
| | | M. | 49.06 ± 3.62 | 95.91 ± 77.52 | **64.73 ± 79.34** | 5.49 ± 2.31 | 11.69 ± 9.71 | **6.12 ± 7.68** |
| spectrum (2) | ⋮⋮ | N. | **167.89 ± 6.55** | 11.42 ± 18.04 | 30.39 ± 17.79 | **18.61 ± 3.55** | 1.26 ± 2.31 | 3.43 ± 2.69 |
| | | B. | 66.23 ± 4.70 | **39.08 ± 63.89** | 104.39 ± 61.07 | 7.41 ± 2.74 | **3.55 ± 5.81** | 12.34 ± 7.66 |
| | | M. | 51.48 ± 5.14 | 41.26 ± 67.57 | **116.95 ± 68.32** | 5.72 ± 2.38 | 5.44 ± 9.12 | **12.15 ± 7.49** |
| Nasci. et al. [10] | Bio. 3.7 | N. | **209.70 ± 4.08** | 0.00 ± 0.00 | 0.00 ± 0.00 | **19.81 ± 3.82** | 2.50 ± 1.53 | 0.99 ± 1.03 |
| | | B. | 0.11 ± 0.47 | **207.64 ± 5.74** | 1.95 ± 4.04 | 0.90 ± 1.12 | **21.47 ± 3.97** | 0.94 ± 1.55 |
| | | M. | 0.05 ± 0.39 | 2.13 ± 4.30 | **207.52 ± 5.77** | 0.75 ± 0.91 | 1.28 ± 1.71 | **21.27 ± 3.91** |
| | Db. 8 | N. | **209.70 ± 4.08** | 0.00 ± 0.00 | 0.00 ± 0.00 | **20.71 ± 3.79** | 2.04 ± 1.53 | 0.55 ± 0.87 |
| | | B. | 0.20 ± 1.07 | **205.76 ± 7.26** | 3.74 ± 5.77 | 1.09 ± 1.14 | **21.07 ± 3.73** | 1.14 ± 1.53 |
| | | M. | 0.06 ± 0.39 | 4.38 ± 6.53 | **205.26 ± 7.69** | 0.41 ± 0.70 | 1.68 ± 2.04 | **21.21 ± 3.81** |
| | Sym. 8 | N. | **209.70 ± 4.08** | 0.00 ± 0.00 | 0.00 ± 0.00 | **20.75 ± 3.78** | 2.06 ± 1.59 | 0.49 ± 0.82 |
| | | B. | 0.72 ± 2.03 | **201.53 ± 8.78** | 7.45 ± 7.45 | 1.16 ± 1.34 | **20.66 ± 3.76** | 1.48 ± 1.63 |
| | | M. | 0.17 ± 0.58 | 6.62 ± 6.62 | **202.91 ± 7.40** | 0.46 ± 0.69 | 1.93 ± 1.71 | **20.91 ± 3.67** |

**Table 5.** Confusion matrix of proposed model and state-of-the-art approaches as feature extraction and SVM neural network with polynomial kernel as classifier. N., B., and M. are Normal, Benign, and Malignant abbreviations, respectively.

| *Feature extraction* | *Function* | | *Train* | | | *Test* | | |
|---|---|---|---|---|---|---|---|---|
| | | | N. | B. | M. | N. | B. | M. |
| Proposed model | Bio. 3.7 | N. | **209.70 ± 4.08** | 0.00 ± 0.00 | 0.00 ± 0.00 | **21.13 ± 3.83** | 1.72 ± 1.22 | 0.45 ± 0.75 |
| | | B. | 0.00 ± 0.00 | **207.55 ± 5.26** | 2.15 ± 3.23 | 0.76 ± 0.93 | **21.25 ± 3.90** | 1.29 ± 1.91 |
| | | M. | 0.00 ± 0.00 | 3.67 ± 3.30 | **206.03 ± 5.22** | 0.56 ± 0.81 | 2.37 ± 2.30 | **20.37 ± 3.91** |
| | Db. 8 | N. | **209.70 ± 4.08** | 0.00 ± 0.00 | 0.00 ± 0.00 | **21.46 ± 3.94** | 1.16 ± 1.07 | 0.68 ± 0.84 |
| | | B. | 0.00 ± 0.00 | **207.18 ± 5.49** | 2.52 ± 3.45 | 0.75 ± 0.94 | **21.16 ± 3.91** | 1.39 ± 2.04 |
| | | M. | 0.00 ± 0.00 | 3.49 ± 3.48 | **206.21 ± 5.31** | 0.47 ± 0.73 | 2.10 ± 2.16 | **20.73 ± 3.91** |
| | Sym. 8 | N. | **209.70 ± 4.08** | 0.00 ± 0.00 | 0.00 ± 0.00 | **21.86 ± 3.94** | 1.04 ± 0.98 | 0.40 ± 0.67 |
| | | B. | 0.00 ± 0.00 | **207.13 ± 5.25** | 2.57 ± 3.39 | 0.70 ± 0.92 | **21.29 ± 3.90** | 1.31 ± 1.86 |
| | | M. | 0.00 ± 0.00 | 3.48 ± 3.76 | **206.22 ± 5.41** | 0.49 ± 0.76 | 2.00 ± 2.05 | **20.82 ± 3.97** |
| spectrum (1) | | | N. | **169.95 ± 7.48** | 23.25 ± 16.84 | 16.49 ± 14.84 | **18.76 ± 3.63** | 2.66 ± 2.40 | 1.88 ± 2.20 |
| | | B. | 36.12 ± 3.24 | **92.39 ± 84.59** | 81.19 ± 81.72 | 4.03 ± 2.07 | **9.09 ± 8.49** | 10.19 ± 10.62 |
| | | M. | 23.62 ± 2.57 | 93.28 ± 85.50 | **92.80 ± 87.93** | 2.71 ± 1.60 | 11.55 ± 10.83 | **9.03 ± 8.54** |
| spectrum (2) | ⋮⋮ | N. | **155.28 ± 5.88** | 34.86 ± 26.06 | 19.56 ± 24.27 | **17.21 ± 3.52** | 3.89 ± 3.39 | 2.21 ± 3.04 |
| | | B. | 24.36 ± 5.00 | **114.42 ± 87.85** | 70.92 ± 84.91 | 2.79 ± 1.83 | **12.26 ± 9.97** | 8.25 ± 10.12 |
| | | M. | 9.23 ± 2.88 | 121.08 ± 95.53 | **79.39 ± 95.30** | 1.01 ± 1.04 | 13.98 ± 11.37 | **8.30 ± 10.21** |
| Nasci. et al. [10] | Bio. 3.7 | N. | **209.70 ± 4.08** | 0.00 ± 0.00 | 0.00 ± 0.00 | **20.22 ± 3.99** | 2.36 ± 1.59 | 0.72 ± 0.89 |
| | | B. | 0.32 ± 1.06 | **188.48 ± 14.38** | 20.90 ± 14.24 | 0.81 ± 1.07 | **19.46 ± 3.97** | 3.02 ± 2.27 |
| | | M. | 0.41 ± 1.01 | 22.78 ± 15.10 | **186.52 ± 15.16** | 0.55 ± 0.78 | 3.33 ± 2.46 | **19.42 ± 3.96** |
| | Db. 8 | N. | **209.70 ± 4.08** | 0.00 ± 0.00 | 0.00 ± 0.00 | **21.08 ± 3.87** | 1.78 ± 1.35 | 0.44 ± 0.69 |
| | | B. | 0.44 ± 1.37 | **184.52 ± 16.69** | 24.74 ± 15.80 | 0.99 ± 1.11 | **18.75 ± 3.99** | 3.56 ± 2.76 |
| | | M. | 0.35 ± 0.92 | 27.06 ± 15.46 | **182.29 ± 15.84** | 0.45 ± 0.73 | 3.90 ± 2.49 | **18.95 ± 3.89** |
| | Sym. 8 | N. | **209.70 ± 4.08** | 0.00 ± 0.00 | 0.00 ± 0.00 | **21.38 ± 3.78** | 1.38 ± 1.24 | 0.54 ± 0.80 |
| | | B. | 0.64 ± 1.54 | **183.29 ± 16.86** | 25.77 ± 16.18 | 1.11 ± 1.25 | **18.66 ± 3.80** | 3.53 ± 2.36 |
| | | M. | 0.22 ± 0.69 | 24.82 ± 14.62 | **184.66 ± 14.72** | 0.45 ± 0.69 | 3.41 ± 2.13 | **19.44 ± 3.96** |

**Table 6.** Student's *t*-test between our proposal, using Symlet 8 wavelets, and the state-of-the-art approaches.

| Classifier | Comparison | *Hypothesis* test | *p*-value |
|---|---|---|---|
| SVM, linear | Proposed model Symlet 8 *vs* Morphological spectrum (1) | 1 | 1.1429e-273 |
|  | Proposed model Symlet 8 *vs* Morphological spectrum (2) | 1 | 2.18377e-265 |
|  | Proposed model Symlet 8 *vs* Nascimento *et al* [10], Biorthogonal 3.7 | 1 | 3.06729e-56 |
|  | Proposed model Symlet 8 *vs* Nascimento *et al* [10], Daubechies 8 | 1 | 7.06147e-42 |
|  | Proposed model Symlet 8 *vs* Nascimento *et al* [10], Symlet 8 | 1 | 3.82393e-55 |
| SVM, polynomial | Proposed model Symlet 8 *vs* Morphological spectrum (1) | 1 | 1.45094e-260 |
|  | Proposed model Symlet 8 *vs* Morphological spectrum (2) | 1 | 1.92266e-228 |
|  | Proposed model Symlet 8 *vs* Nascimento *et al* [10], Biorthogonal 3.7 | 1 | 1.09343e-67 |
|  | Proposed model Symlet 8 *vs* Nascimento *et al* [10], Daubechies 8 | 1 | 2.86154e-66 |
|  | Proposed model Symlet 8 *vs* Nascimento *et al* [10], Symlet 8 | 1 | 4.71656e-64 |
| SVM, RBF | Proposed model Symlet 8 *vs* Morphological spectrum (1) | 1 | 3.21756e-37 |
|  | Proposed model Symlet 8 *vs* Morphological spectrum (2) | 1 | 2.93787e-26 |
|  | Proposed model Symlet 8 *vs* Nascimento *et al* [10], Biorthogonal 3.7 | 1 | 7.59836e-114 |
|  | Proposed model Symlet 8 *vs* Nascimento *et al* [10], Daubechies 8 | 1 | 2.89142e-114 |
|  | Proposed model Symlet 8 *vs* Nascimento *et al* [10], Symlet 8 | 1 | 9.87814e-114 |
| Decision tree | Proposed model Symlet 8 *vs* Morphological spectrum (1) | 1 | 3.32378e-99 |
|  | Proposed model Symlet 8 *vs* Morphological spectrum (2) | 1 | 1.52165e-52 |
|  | Proposed model Symlet 8 *vs* Nascimento *et al* [10], Biorthogonal 3.7 | 1 | 6.77493e-05 |
|  | Proposed model Symlet 8 *vs* Nascimento *et al* [10], Daubechies 8 | 1 | 3.2631e-10 |
|  | Proposed model Symlet 8 *vs* Nascimento *et al* [10], Symlet 8 | 1 | 1.1476e-19 |
| K-NN | Proposed model Symlet 8 *vs* Morphological spectrum (1) | 1 | 7.92202e-137 |
|  | Proposed model Symlet 8 *vs* Morphological spectrum (2) | 1 | 1.05494e-139 |
|  | Proposed model Symlet 8 *vs* Nascimento *et al* [10], Biorthogonal 3.7 | 1 | 1.21951e-21 |
|  | Proposed model Symlet 8 *vs* Nascimento *et al* [10], Daubechies 8 | 1 | 2.42024e-05 |
|  | Proposed model Symlet 8 *vs* Nascimento *et al* [10], Symlet 8 | 1 | 0.00103431 |

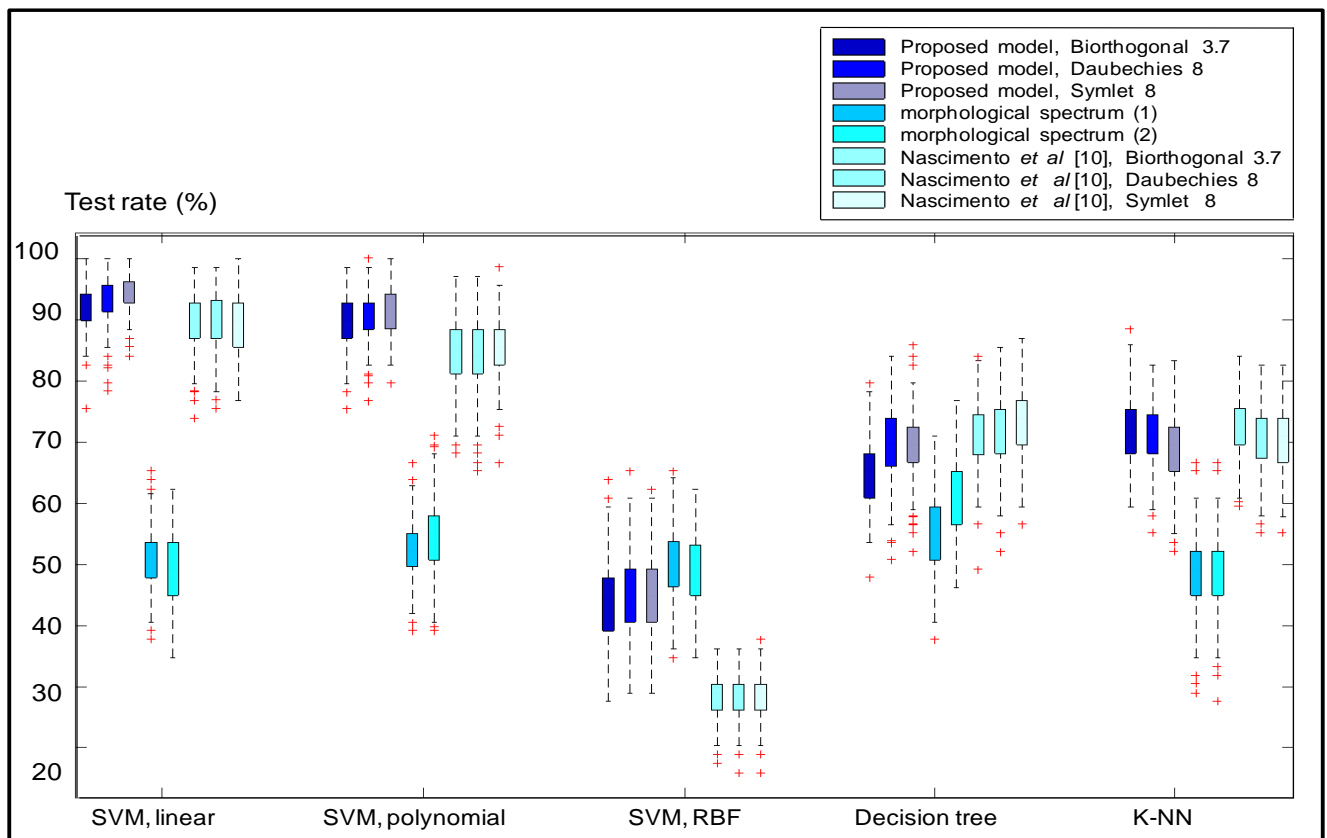

**Fig. 9. Boxplots of the percentage accuracy rates obtained with our proposals and the state-of-the-art techniques.**

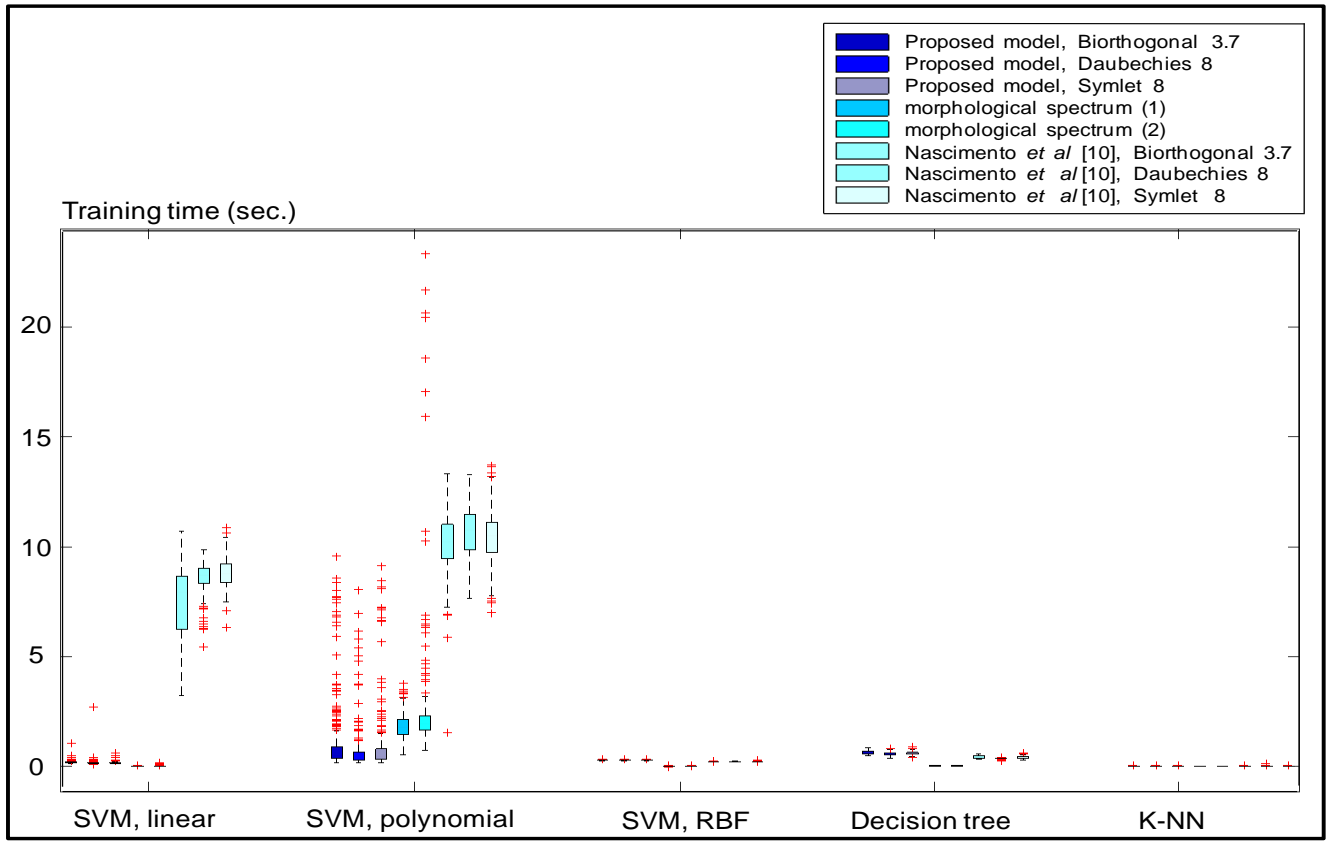

**Fig. 10.** Boxplots of the training times, in seconds, obtained with our proposals and the state-of-the-art techniques.

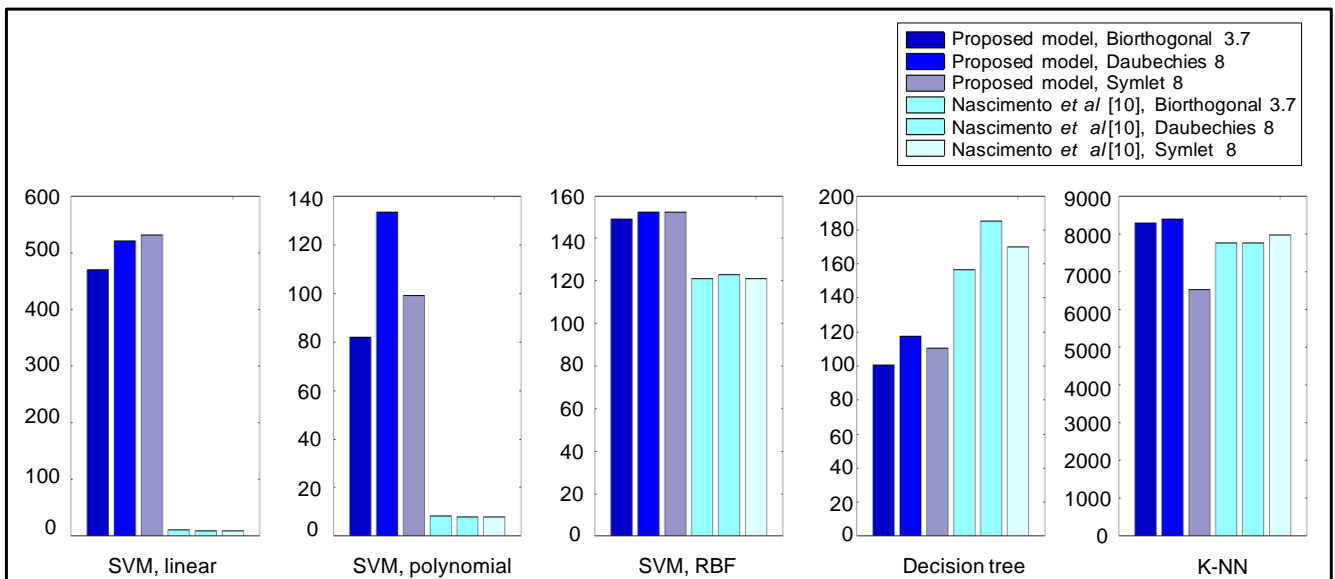

**Fig. 11.** Ratios between medium percentage accuracy rate and medium training time for each studied method.

In order to facilitate comparisons among all methods, considering both accuracy rates and training times, we defined the ration between average percentage accuracy and average training time. Fig. 11 displays these ratios for the classifiers we evaluated. It is noteworthy that the relationship is in a inverse order.

The morphological spectrum applications showed the low training times compared to other approaches, in all kernels. The accuracy, however, was much lower than the other techniques, in all tests. Accuracy rates are around 50%. Therefore, the use of morphological spectra as feature extractors was not able to get feasible results in this application. Although this approach could achieve the highest ratios between average accuracy and training time, it does not make sense adopting a technique with accuracy rates around 50%, because it is almost a random result. This is the reason why we excluded results acquired with morphological spectra from the general results presented on Fig. 11.

From Fig. 11, considering linear kernels, our proposal demonstrated a clear large advantage when compared to the work of Nascimento et al. (2013). Using the wavelet function Symlet 8 our proposal was able to achieve the best average accuracy with 94.11% with corresponding training time of 0.18 seconds. Nascimento et al. (2013) with wavelet family Daubechies 8 achieve their best average accuracy with 90.12% with corresponding training time of 8.61 seconds. In this case, the ratio between average accuracy and training time obtained with our proposal was 50 times higher than the ratio obtained using the best method of the state-of-the-art.

Taking into account polynomial kernels, our proposal still take large advantages over the works of the state-of-the-art, but with a smaller margin. Considering RBF kernels, our model reached slightly higher results than Nascimento et al. (2013).

Even regarding to Fig. 11, in relation to Decision Tree classifier, Nascimento et al. (2013) achieved results better than ours. While on K-NN classifier, our results were slightly higher with expect of wavelet family Symlet 8.

Our proposal could achieve the best results employing feature extraction with wavelets Symlet 8 and SVM networks with linear kernels, reaching an average accuracy of 94.11%. This is the reason to adopt this configuration as a reference to be employed other classifiers. We explore MLP, SVM, and ELM neural networks. Besides we use two other classifiers employed by state-of-the-art; Decision Tree and K-NN.

Table 7 shows the classification results using MLP networks. MLP classifiers have high training times, with high standard deviations as well. This indicates that MLP networks are sensibly affected by abrupt changes in relation to training time, depending on the initial conditions.

Taking into account overall accuracy, MLP classifiers got the minimum accuracy rate was just 37.68% using batch training learning rules. This approach has two hidden layers, each one with 100 neurons. The observed maximum accuracy was 89.61% for Conjugate gradient with Fletcher-Reeve, it employed only one hidden layer architecture with 100 neurons.

**Table 7. Percentage accuracies in training and test stages, training and test time for different MLP-based classifiers, for our proposal using Symlet 8 wavelets.**

|  | *Train rate (%)* | *Test rate (%)* | *Train time (sec.)* | *Test time (sec.)* |
|---|---|---|---|---|
| Batch training learning rules | 43.06 ± 18.55 | 41.84 ± 16.15 | 7.86 ± 21.19 | **0.05 ± 0.03** |
| Batch training learning rules * | 39.37 ± 11.55 | 38.82 ± 11.44 | 3.17 ± 12.90 | 0.06 ± 0.02 |
| Batch training learning rules ** | 38.01 ± 10.57 | 37.68 ± 10.93 | 3.51 ± 17.18 | 0.11 ± 0.03 |
| Conjugate gradient with Powell-Beale restarts | 95.88 ± 5.85 | 88.03 ± 6.93 | 6.18 ± 2.87 | **0.05 ± 0.03** |
| Conjugate gradient with Powell-Beale restarts * | 93.68 ± 6.70 | 85.10 ± 7.38 | 7.11 ± 3.10 | 0.06 ± 0.03 |
| Conjugate gradient with Powell-Beale restarts ** | 97.01 ± 5.13 | 87.46 ± 5.81 | 23.49 ± 8.01 | 0.11 ± 0.03 |
| Conjugate gradient with Fletcher-Reeves | **97.74 ± 3.37** | **89.61 ± 5.34** | 9.09 ± 5.15 | **0.05 ± 0.03** |
| Conjugate gradient with Fletcher-Reeves * | 96.22 ± 4.85 | 87.30 ± 6.56 | 11.26 ± 7.58 | 0.06 ± 0.03 |
| Conjugate gradient with Fletcher-Reeves ** | 96.93 ± 6.02 | 87.29 ± 6.41 | 37.99 ± 21.59 | 0.10 ± 0.03 |
| Conjugate gradient with Polak-Ribiere | 91.58 ± 7.85 | 83.49 ± 7.85 | 4.69 ± 2.42 | **0.05 ± 0.03** |
| Conjugate gradient with Polak-Ribiere * | 90.36 ± 7.47 | 81.83 ± 7.31 | 6.06 ± 2.75 | 0.07 ± 0.03 |
| Conjugate gradient with Polak-Ribiere ** | 89.75 ± 8.96 | 80.49 ± 8.38 | 18.67 ± 8.41 | 0.11 ± 0.03 |
| Gradient descent backpropagation | 96.91 ± 1.05 | 85.64 ± 3.53 | 33.82 ± 2.37 | **0.05 ± 0.03** |
| Gradient descent backpropagation * | 94.52 ± 3.13 | 83.30 ± 4.66 | 37.92 ± 9.57 | 0.07 ± 0.03 |
| Gradient descent backpropagation ** | 82.15 ± 22.34 | 73.17 ± 18.07 | 108.05 ± 65.40 | 0.11 ± 0.03 |
| Gradient descent with momentum | 43.08 ± 17.98 | 41.87 ± 16.10 | 3.37 ± 9.26 | 0.06 ± 0.03 |
| Gradient descent with momentum * | 42.92 ± 10.33 | 42.33 ± 10.71 | 1.61 ± 5.80 | 0.07 ± 0.03 |
| Gradient descent with momentum ** | 41.37 ± 15.80 | 40.28 ± 14.26 | 8.99 ± 32.08 | 0.11 ± 0.03 |
| Gradient descent with adaptive learning | 64.35 ± 8.35 | 60.81 ± 7.77 | 2.18 ± 1.29 | **0.05 ± 0.03** |
| Gradient descent with adaptive learning * | 66.28 ± 7.52 | 62.42 ± 7.46 | 2.85 ± 1.43 | 0.06 ± 0.03 |
| Gradient descent with adaptive learning ** | 58.72 ± 6.30 | 56.57 ± 6.76 | 7.64 ± 4.98 | 0.11 ± 0.03 |
| Gradient descent combined | 80.61 ± 20.98 | 71.78 ± 17.02 | 3.01 ± 1.21 | **0.05 ± 0.03** |
| Gradient descent combined * | 76.84 ± 21.39 | 69.29 ± 17.31 | 3.74 ± 1.79 | 0.07 ± 0.03 |
| Gradient descent combined ** | 78.24 ± 17.86 | 68.24 ± 13.43 | 11.36 ± 4.91 | 0.11 ± 0.03 |
| One-step secant backpropagation | 94.17 ± 6.18 | 85.37 ± 6.41 | 5.54 ± 2.25 | **0.05 ± 0.03** |
| One-step secant backpropagation * | 92.34 ± 7.03 | 83.23 ± 7.27 | 7.27 ± 3.04 | 0.07 ± 0.03 |
| One-step secant backpropagation ** | 92.06 ± 9.77 | 81.82 ± 8.72 | 24.15 ± 9.78 | 0.11 ± 0.03 |
| Resilient backpropagation (Rprop) | 69.20 ± 17.55 | 62.35 ± 14.26 | 1.00 ± 0.33 | **0.05 ± 0.03** |
| Resilient backpropagation (Rprop) * | 47.52 ± 19.41 | 44.91 ± 16.75 | **0.98 ± 0.42** | 0.07 ± 0.03 |
| Resilient backpropagation (Rprop) ** | 37.36 ± 13.56 | 36.37 ± 12.56 | 1.09 ± 1.09 | 0.12 ± 0.03 |
| Scaled conjugate gradient backpropagation | 94.51 ± 5.69 | 85.76 ± 6.68 | 3.16 ± 1.23 | **0.05 ± 0.03** |
| Scaled conjugate gradient backpropagation * | 93.08 ± 6.34 | 83.98 ± 6.86 | 4.10 ± 1.56 | 0.06 ± 0.03 |
| Scaled conjugate gradient backpropagation ** | 91.57 ± 11.15 | 81.37 ± 10.09 | 12.61 ± 4.67 | 0.11 ± 0.03 |

Table 8 shows results of the ELMs neural networks in a multi-kernel approach. ELM networks achieved a maximum accuracy rate of 83.58% for linear kernels. Training time was considerably short. For linear and polynomial kernels, ELM classifiers could reach high performances.

Table 9 displays results obtained using SVM classifiers. Taking into account accuracy rates, classifiers based on linear and polynomial kernels obtained superior performance to other classifiers. The configuration based on linear kernels achieved the best accuracy of all networks, getting 94.11% in distinction of normal, benign and malignant cases. Regarding the training time, SVM networks are not as fast as ELMs. The learning process, however, consumes so much less time compared with MLP networks in general.

In Table 10, Decision Tree was just slightly higher than K-NN. They achieved performances of 69.26% and 68.95%, respectively. K-NN, however, consumes so much less time in training process. Training time of K-NN classifier is comparable to ELM networks

**Table 8. Results of feature extraction of proposed model by wavelets, with Symlet 8 function. The classifiers are ELMs networks for grouping into 3 classes: normal, benign and malignant.**

| | *Train rate (%)* | *Test rate (%)* | *Train time (sec.)* | *Test time (sec.)* |
|---|---|---|---|---|
| ELM, RBF | 99.99 ± 0.04 | 43.07 ± 6.06 | 0.09 ± 0.03 | 0.02 ± 0.04 |
| ELM, linear | **100.00 ± 0.02** | **83.58 ± 4.34** | 0.05 ± 0.03 | 0.01 ± 0.02 |
| ELM, polynomial | 99.36 ± 0.50 | 82.68 ± 4.37 | 0.06 ± 0.03 | 0.01 ± 0.02 |
| ELM, wavelet | 99.99 ± 0.03 | 38.39 ± 5.86 | 0.14 ± 0.03 | 0.02 ± 0.04 |
| ELM, sigmoid | 34.24 ± 0.43 | 28.02 ± 3.32 | 0.03 ± 0.03 | 0.00 ± 0.02 |
| ELM, sine | 56.04 ± 1.78 | 33.39 ± 5.61 | 0.04 ± 0.03 | **0.00 ± 0.01** |
| ELM, hard limit | 33.96 ± 0.33 | 27.73 ± 2.92 | 0.03 ± 0.03 | 0.00 ± 0.02 |
| ELM, tribas | 33.36 ± 0.59 | 33.33 ± 5.26 | **0.02 ± 0.03** | **0.00 ± 0.01** |

**Table 9. Results of feature extraction of proposed model by wavelets, with Symlet 8 function. The classifiers are SVMs networks for grouping into 3 classes: normal, benign and malignant.**

| | *Train rate (%)* | *Test rate (%)* | *Train time (sec.)* | *Test time (sec.)* |
|---|---|---|---|---|
| SVM, linear kernel | **99.80 ± 0.18** | **94.11 ± 3.00** | **0.18 ± 0.05** | **0.01 ± 0.01** |
| SVM, polynomial kernel | 99.04 ± 0.49 | 91.51 ± 3.43 | 0.92 ± 1.44 | **0.01 ± 0.01** |
| SVM, RBF kernel | 98.81 ± 0.38 | 44.63 ± 6.49 | 0.29 ± 0.01 | 0.04 ± 0.01 |
| SVM, sigmoid kernel | 31.43 ± 4.72 | 27.03 ± 5.28 | 0.28 ± 0.01 | 0.04 ± 0.01 |

**Table 10. Results of feature extraction of proposed model by wavelets, with Symlet 8 function. The classifiers are Decision tree and K-NN networks for grouping into 3 classes: normal, benign and malignant.**

| | *Train rate (%)* | *Test rate (%)* | *Train time (sec.)* | *Test time (sec.)* |
|---|---|---|---|---|
| Decision tree | 97.13 ± 0.66 | **69.26 ± 5.39** | 0.63 ± 0.06 | **0.00 ± 0.00** |
| K-NN | **100.00 ± 0.00** | 68.95 ± 5.29 | **0.01 ± 0.01** | 0.02 ± 0.01 |

## 7. Conclusion

Herein this work we proposed a methodology to deal with mammographic image classification using feature extraction based on the calculation of Zernike moments from a series of multiresolution image components, obtained by the series of wavelets; and classification based on kernel-oriented neural networks, specifically support vector machines and extreme learning machines, varying the kernels of the neurons of the hidden layers.

Taking into account BIRADS criteria, we defined the classes of interest as normal (mammary tissue without lesion), benign, and malignant lesion, in an effort to build an adequate data representation to simultaneously promote detection and classification of mammary lesions. The theoretical basis of Mathematical Morphology affirms pattern spectra as ideal and unique shape representations of binary images. Due to this aspect, we also performed feature extraction experiments using morphological spectra and evaluated them, though IRMA's image database is composed by gray-level mammographic images.

We generated results exploring the relatively large amount of wavelet families. The best results were obtained using multiresolution wavelets Symlet 8 at feature extraction. In order to compare our best results with classical learning-based approaches, we also performed classification experiments with multilayer perceptrons whose training processes were guided by eleven different algorithms.

In order to get more precise performance comparisons, integrating training time and classification accuracy, we stablished a ration between overall accuracy and training time. Our proposal, with multiresolution wavelets, Zernike moments, and kernel-based learning machines with linear kernels, resulted much more efficient than the proposals of the state-of-the-art based on wavelets, energy moments and kernel-based machines with the same kernels, taking into account the ration between overall accuracy and training time. This advantage was reduced when we used polynomial kernels. However, the situation changes when we used kernel-based learning machines with RBF kernels: the proposals of the state-of-the-art reached superior results, but not much more higher than our proposals, as when we used linear kernels.

The proposed methods and methodology can be extent to deal with more difficult mammographic images, like dense, extremely dense, and fibroglandular mammary tissues. Our proposal could also be applied to other biomedical image classification problems related to the diagnosis of breast cancer and other forms of cancer, like mammary and brain magnetic resonance imaging, immunohistochemistry images, and other complex image analysis problems in which both shape and texture are essential aspects for classification, and high overall accuracies combined with relatively low training times are needed in order to promote accurate diagnoses as fast as necessary.

**Acknowledgements**

The authors thank Brazil's REUNI (ref. 01/2012) for the partial financial support. The IRMA database is courtesy of Prof. Thomas M. Deserno, from the Department of Medical Informatics at RWTH Aachen University, Germany.